\documentclass[conference,11pt]{IEEEtran}
\IEEEoverridecommandlockouts
\usepackage{cite}
\usepackage{amsmath,amssymb,amsfonts}
\usepackage{graphicx}
\usepackage{amsmath}
\usepackage{algorithm}
\usepackage{algpseudocode}
\usepackage{algorithmicx}
\usepackage{textcomp}
\usepackage{booktabs}  
\usepackage{enumitem}  
\usepackage{xcolor}
\usepackage{caption}
\usepackage{geometry}
\usepackage{tikz}
\usepackage{subcaption}
\usetikzlibrary{shapes, arrows.meta,shapes.geometric, arrows, positioning, calc}
\usepackage{pgfplots}
\pgfplotsset{compat=1.18}
\usepackage{booktabs}
\usepackage{multirow}
\usepackage{lmodern}
\pgfplotsset{compat=newest}
\usetikzlibrary{plotmarks}
\usepackage{textcomp}

\algnewcommand{\LineComment}[1]{\State \(\triangleright\) #1}
\def\BibTeX{{\rm B\kern-.05em{\sc i\kern-.025em b}\kern-.08em
    T\kern-.1667em\lower.7ex\hbox{E}\kern-.125emX}}
\begin{document}

\title{Trust Aware Federated Learning for Secure Bone Healing Stage Interpretation in e-Health\\
}

\author{
\IEEEauthorblockN{\textbf{Paul Shepherd}\textsuperscript{1,2}, \textbf{Tasos Dagiuklas}\textsuperscript{1}, \textbf{Bugra Alkan}\textsuperscript{1}}
\IEEEauthorblockN{\textbf{Joaquim Bastos}\textsuperscript{2}, \textbf{Jonathan Rodriguez}\textsuperscript{2}}
\IEEEauthorblockA{
\textsuperscript{1}London South Bank University, London, UK \\
\textsuperscript{2}Instituto de Telecomunicações, Aveiro, Portugal \\
paul@av.it.pt, tdagiuklas@lsbu.ac.uk, alkanb@lsbu.ac.uk, jbastos@av.it.pt, jonathan@av.it.pt
}
}
\maketitle
\begin{abstract}
This paper presents a trust aware federated learning (FL) framework for interpreting bone healing stages using spectral features derived from frequency response data. The primary objective is to address the challenge posed by either unreliable or adversarial participants in distributed medical sensing environments. The framework employs a multi-layer perceptron model trained across simulated clients using the Flower FL framework. The proposed approach integrates an Adaptive Trust Score Scaling and Filtering (ATSSSF) mechanism with exponential moving average (EMA) smoothing to assess, validate and filter client contributions. 

Two trust score smoothing strategies have been investigated, one with a fixed factor and another that adapts according to trust score variability. Clients with low trust are excluded from aggregation and readmitted once their reliability improves, ensuring model integrity while maintaining inclusivity. Standard classification metrics have been used to compare the performance of ATSSSF with the baseline Federated Averaging strategy. Experimental results demonstrate that adaptive trust management can improve both training stability and predictive performance by mitigating the negative effects of compromised clients while retaining robust detection capabilities. The work establishes the feasibility for adaptive trust mechanisms in federated medical sensing and identifies extension to clinical cross silo aggregation as a future research direction.
\end{abstract}


\section{Introduction}
Federated learning (FL) enables distributed model training while preserving data privacy, which is particularly important in healthcare where restrictions prevent centralizing patient records [1,2]. Despite this advantage, FL is exposed to challenges [11] such as data heterogeneity and either unreliable or adversarial participants that may reduce overall model quality [6, 7].

Bone healing stage classification represents a relevant application domain, as conventional assessment methods rely primarily on radiographic imaging or mechanical load testing, both of which are resource-intensive, intermittently applied, and limited in their ability to provide continuous, non-invasive monitoring of healing progression. Frequency response based spectral features [3] provide a non invasive alternative, and FL allows institutions (cross silo) to collaboratively learn from these data without disclosing raw signals.

In such collaborative setups, trust management (TM) is required to safeguard aggregation from client updates that are noisy, corrupted, or adversarial [13, 14, 15]. Standard algorithms like FedAvg [5,7] do not incorporate client reliability, while existing robust aggregation methods often rely on either fixed rules or predefined attack assumptions, reducing their flexibility in real in situ deployments.

To address this gap, this paper applies the ATSSSF [4, 6] mechanism to the task of bone healing stage interpretation. ATSSSF evaluates client weight contributions using multiple performance metrics combined through the TOPSIS method [16, 17, 23] and incorporates temporal smoothing via EMA [18]. Both static and adaptive smoothing strategies are examined to assess their effect on preserving model reliability and deployment.

This paper introduces a trust regulated FL framework for bone healing stage classification using spectral sensing features, where ATSSSF based client evaluation is combined with variance aware EMA smoothing. The study demonstrates how adaptive trust dynamics improve aggregation stability and predictive consistency relative to baseline FedAvg [21].
\section{Related Work and Background}
FL is now central to privacy preserving collaboration in healthcare, particularly for medical imaging, electronic health records, and wearable sensor data [1,2]. These models, such as those based on FedAvg, enable cross institutional training without centralized data, However, FL is inherently challenged by non-IID data distributions and the presence of noisy or malicious client updates, which may destabilise aggregation and reduce predictive reliability. FL has revolutionized applications in disease prediction and image classification, yet its use for biomechanical or physiological time-series signals, like those needed for bone healing assessment—remains uncommon [1]. Most sensor based FL studies focus on remote patient monitoring rather than orthopedic frequency response data, making this application area largely unexplored.

Robustness in FL traditionally relies on aggregation algorithms such as Krum and FLCert [12] to limit the effect of outliers, with hierarchical and clustering approaches tackling non-IID data and scalability [9, 20]. Bayesian aggregation introduces uncertainty aware weighting of client updates, but typically incurs higher computational overhead and relies on predefined adversarial assumptions, limiting flexibility in dynamic healthcare FL environments [1].

Trust based client scoring is emerging as a robust filtering mechanism, evaluating client contributions using indicators such as local validation accuracy, gradient consistency, update divergence, or label reliability [5]. However, most existing approaches rely on static trust thresholds and fixed smoothing parameters, limiting adaptability to evolving participation patterns and non stationary data distributions. Recent work suggests that adaptive trust mechanisms using methods like variable smoothing or dynamic thresholds could provide the flexibility needed in healthcare FL [4], but such techniques are just starting to be explored in FL for sensor modalities. Furthermore, FL for bone healing monitoring, using frequency response data remains a novel and underexplored area [2]. Adaptive EMA smoothing, which adjusts to trust score variance, is uncommon, despite its potential to improve reliability in medical sensing settings where stable interpretation is vital [4].

Unlike parameter level robust aggregation methods such as Krum or Trimmed Mean, which rely on distance-based filtering under explicit Byzantine assumptions, ATSSSF performs client level trust assessment using multi-criteria performance indicators. This enables dynamic exclusion and re-admission without assuming a fixed adversarial fraction. While robust aggregators operate at the update level under predefined attack models, ATSSSF regulates participation at the client level and is orthogonal to such defences, allowing potential integration within the same framework.

This work introduces a TOPSIS based adaptive trust framework for federated bone healing stage detection, integrating a variable EMA that adjusts its smoothing coefficient according to trust score dynamics [4]. This approach aims to maintain reliable performance and support effective decision making under varying participation and data conditions, moving beyond static robustness methods in medical FL.
\section{Methodology and Framework}
The dataset comprises spectral measurements corresponding to seven bone healing stages, ranging from Fresh Fracture to Fully Healed. For each sample, the frequency domain scattering parameters $S_{11}(f)$ and $S_{21}(f)$ were recorded across a defined frequency sweep. Derived features, including the $S_{11}-S_{21}$ difference, the $S_{11}/S_{21}$ ratio, and first order spectral derivatives, were computed to enhance class separability. Data were normalised using standard scaling, and class imbalance in the training set was addressed using the Synthetic Minority Oversampling Technique (SMOTE), while the test set remained unchanged. Healing stage labels were encoded categorically.

The dataset reflects controlled spectral acquisitions rather than multi patient data, and therefore lacks natural subject variability. To emulate distributed conditions, the data have been partitioned and synthetically diversified to simulate heterogeneous and potentially unreliable client behavior. This setup provides a controlled environment for evaluating the proposed TM framework within a FL scenario.

A feedforward neural network has been used for stage classification. The model comprised three dense layers with ReLU activation, dropout, and batch normalization. The output layer has applied a softmax function across seven classes. Training has used categorical cross entropy loss and the Adam optimizer with class weight balancing. Model evaluation has employed accuracy, macro precision, macro-recall, and macro F1, ensuring balanced assessment across stages.

The FL environment has been implemented using the Flower framework [22]. The dataset has been divided across simulated clients, each performing local updates for several rounds. Global aggregation has followed the FedAvg baseline, computing a weighted mean of client parameters while maintaining data locality and privacy.

To enhance robustness, ATSSSF mechanism [4] has been incorporated into aggregation. Client trust has been computed using the TOPSIS method, combining multiple criteria local validation accuracy, contribution similarity, and update consistency into a normalized trust score between 0 and 1. Clients below the 0.75 trust threshold has been excluded from aggregation. The trust threshold $(\tau = 0.75)$ and the maximum client omission cap $(m = 3)$ have been selected conservatively to balance robustness against unreliable updates with aggregation continuity, consistent with prior ATSSSF study [4]. While fixed in this study to isolate the effect of adaptive trust smoothing, these parameters are not intrinsic to the framework and can be made adaptive in future work.

Trust scores have been stabilized using EMA smoothing. Unlike the static EMA $(\alpha = 0.3)$, this study has employed an adaptive EMA, where the smoothing coefficient adjusted dynamically based on trust score variance. High variance reduced $(\alpha)$ to emphasize stability, while low variance increased $(\alpha)$ to enable faster response to genuine performance shifts. This adaptation improved trust consistency and decision reliability.

Client omission has followed a conservative policy limiting exclusions to three clients per round. Readmission has occurred once a client’s smoothed trust exceeded the threshold for two consecutive rounds, allowing recovery from temporary instability. This policy has preserved both robustness and fairness, ensuring the federated network remained stable under dynamic participation conditions.
\begin{algorithm}[htbp]
\caption{ATSSSF Workflow in FL}
\label{alg:atsssf_fl}
\begin{algorithmic}[1]
\footnotesize
\Require Client set $C$, weights $W$, threshold $\tau = 0.75$, smoothing $\alpha$, max omissions $m = 3$, rounds $T$
\Ensure Global model $w_{global}$

\State Initialize $w_{global}^{(0)}$, $\hat{T}_i^{(0)} \gets 1$ $\forall i$
\For{$t = 1$ to $T$}
    \State Select clients $\mathcal{S}_t \subseteq C$, broadcast $w_{global}^{(t)}$
    \For{each $c_i \in \mathcal{S}_t$}
        \State Train $\rightarrow \Delta w_i^{(t)}$, get metrics $(Acc_i, Prec_i, Rec_i, F1_i)$
        \State $D_i = [Acc_i, Prec_i, Rec_i, F1_i]$
    \EndFor
    
    \State Construct $D = [d_{ij}]_{n \times 4}$
    \For{each metric $j$}
        \State $r_{ij} = \frac{d_{ij}}{\sqrt{\sum_i d_{ij}^2}}$, $v_{ij} = w_j \cdot r_{ij}$
    \EndFor

    \State $A^+ = \{\max_i v_{ij}\}$, $A^- = \{\min_i v_{ij}\}$
    \For{each $c_i$}
        \State $S_i^+ = \sqrt{\sum_j (v_{ij} - A^+_j)^2}$, $S_i^- = \sqrt{\sum_j (v_{ij} - A^-_j)^2}$
        \State $T_i = \frac{S_i^-}{S_i^+ + S_i^-}$
    \EndFor

    \State Compute variance $\sigma^2$ of $T_i$
    \If{$\sigma^2 > \text{threshold}$}
        \State $\alpha \gets 0.5\alpha$ \Comment{Increase smoothing}
    \Else
        \State $\alpha \gets \min(\alpha + 0.05, 1.0)$ \Comment{Faster adaptation}
    \EndIf

    \For{each $c_i$}
        \State $\hat{T}_i^{(t)} = \alpha T_i + (1 - \alpha)\hat{T}_i^{(t-1)}$
        \If{$\hat{T}_i^{(t)} < \tau$}
            \State Omit $c_i$ (max $m$ per round)
        \ElsIf{$\hat{T}_i^{(t)} \ge \tau$ for 2 rounds}
            \State Re-admit $c_i$
        \EndIf
    \EndFor

    \State $w_{global}^{(t+1)} = \sum_{i \in \mathcal{S}_t \setminus O_t} \frac{n_i}{N} \Delta w_i^{(t)}$
\EndFor
\State \Return $w_{global}^{(T)}$
\end{algorithmic}
\end{algorithm}

\begin{figure}[htbp]
    \centering
    \resizebox{3.2in}{!}{ 
    \begin{tikzpicture}[
      node distance=0.9cm and 1.2cm,
      process/.style={rectangle, draw, rounded corners, text centered, text width=3.6cm, minimum height=2.8em, fill=white!40, font=\normalsize},
      adp_process/.style={rectangle, draw=black, thick, rounded corners, text centered, text width=4.0cm, minimum height=3.8em, fill=green!20, font=\normalsize},
      arrow/.style={-{Latex[scale=1.0]}, thick},
      entity/.style={ellipse, draw, fill=white!40, minimum height=2.5em, text centered, font=\bfseries\normalsize},
      omitted/.style={dashed, draw=red, fill=red!20, rounded corners, text centered, text width=3.4cm, minimum height=2.8em, font=\normalsize},
      main/.style={draw, fill=white!30, rounded corners, text centered, minimum height=3.0em, text width=4.2cm, font=\normalsize},
      comment/.style={midway, above, sloped, font=\small, text=black}
      ]

      \node[entity] (clients) {Client Devices / Clinics};
      \node[entity, right=of clients, xshift=6cm] (server) {Federated Server};

      \node[main, below=of clients] (prep) {Spectral Data Collection \\ \& Preprocessing};
      \node[process, below=of prep] (local) {Local Model Training \\ (NN on S-parameters)};
      \node[process, below=of local] (topsis) {TOPSIS Based Trust Score \\ Computation: $T_i = f(\text{Acc}, \text{Prec}, \text{Rec}, F1)$};

      \node[process, below=of topsis] (ema) {Exponential Moving Average (EMA) \\ Static $\alpha = 0.3$};
      \node[adp_process, right=of ema, xshift=3.2cm] (adaptive) {Adaptive EMA and ATSSSF \\ Dynamic $\alpha$ \& $\tau$ Adjustment \\ Based on Variance in $T_i$};

      \node[process, below=of ema] (agg) {FedAvg Aggregation with \\ Trust Aware Filtering};
      \node[process, below=of agg] (update) {Global Model Update};

      \node[omitted, right=of agg, xshift=3.2cm] (omit) {Omitted Clients \\ ($\hat{T}_i < 0.75$, Max 3 per Round)};

      \draw[arrow] (clients) -- (prep);
      \draw[arrow] (prep) -- (local);
      \draw[arrow] (local) -- (topsis);
      \draw[arrow] (topsis) -- (ema);
      \draw[arrow, blue, thick] (ema) -- (adaptive);
      \draw[arrow] (ema) -- (agg);
      \draw[arrow] (adaptive) |- (agg.east);
      \draw[arrow] (agg) -- (update);
      \draw[arrow] (update) -- (server);

      \draw[arrow, dashed, red] (adaptive.south) |- (omit.north);

      \draw[arrow, dashed] (server) -- ++(0,-2.5cm) -| (clients) node[comment, xshift=-1cm, yshift=2.2cm] {Global Model Distribution};
    \end{tikzpicture}
    }
    \caption{FL workflow for bone healing stage prediction with ATSSSF.}
    \label{fig:atsssf_bone_framework}
\end{figure}
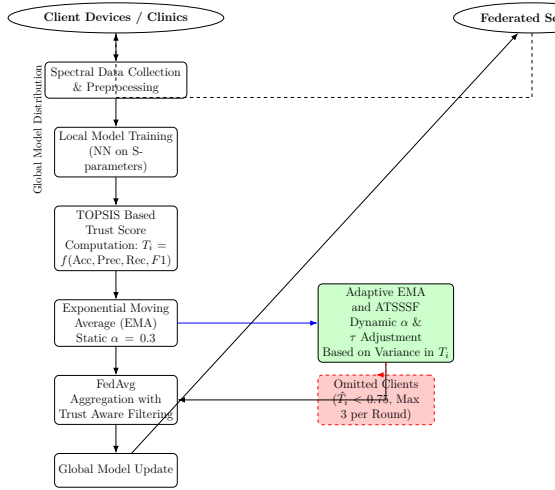
\paragraph{Analytical Insights.}
EMA based trust smoothing acts as a low pass filter on bounded trust scores, reducing temporal variance and stabilising client selection decisions under transient performance fluctuations. Limiting client omission to at most $m$ clients per round guarantees that at least $(N-m)/N$ of the aggregate update mass is retained, thereby preserving aggregation continuity. Adaptive EMA further balances stability and responsiveness by coupling trust reactivity to observed trust variance.
\section{Experimental Setup}
Experiments have been conducted using the cleaned bone healing dataset containing spectral S-parameter features across seven healing stages from Fresh Fracture to Fully Healed. The data have been standardized and split into 80\% training and 20\% testing sets, with class imbalance mitigated using SMOTE on the training portion only. FL was simulated in the Flower framework with 100 virtual clients trained over 500 communication rounds, using the Adam optimizer (learning rate 0.001, batch size 16). The baseline aggregation followed the FedAvg strategy, while the ATSSSF mechanism was integrated to assign and adjust client trust scores derived from local accuracy, precision, recall, and F1 score using the TOPSIS method. Clients with scores below a 0.75 trust threshold were omitted, limited to three per round. EMA smoothing was applied to stabilize temporal trust variations, first with a static coefficient $(\alpha = 0.3)$ and later with an adaptive version that adjusted $(\alpha)$ dynamically based on client variance. 

High test set accuracy, precision, recall, and macro F1 scores have been evaluated to track model behaviour, but these metrics were not the primary objective. Instead, the focus was on demonstrating the proof of concept for robust trust-aware aggregation. Trust score distributions and client omission frequencies have been monitored throughout to assess the stability and resilience of the aggregation process. The evaluation has incorporated test set accuracy, macro precision, recall, and macro F1, as well as monitoring trust score distributions and omission frequencies; however, achieving high performance was not the main objective. Instead, the analysis has served to validate the framework’s feasibility for robust and stable trust aware aggregation in a federated setting.
\section{Results and Analysis}
The dataset consists of controlled spectral acquisitions rather than multi patient clinical data. This choice was intentional, as the aim of this study is to evaluate trust dynamics and aggregation stability in FL, rather than clinical generalisation. Client heterogeneity and unreliability were therefore introduced synthetically to provide a reproducible testbed for trust aware aggregation.

The baseline FedAvg configuration has achieved an overall accuracy of 67.4\%, with a macro F1 score of 0.61 and precision of 0.64 across the seven bone healing stages. Misclassifications were most frequent in early stage classes such as Fresh Fracture and Soft Callus, where spectral overlap reduced separability, while later stages such as Near Healing and Fully Healed showed higher accuracy.
\begin{figure}[htbp]
\centering
\includegraphics[width=1.0\columnwidth]{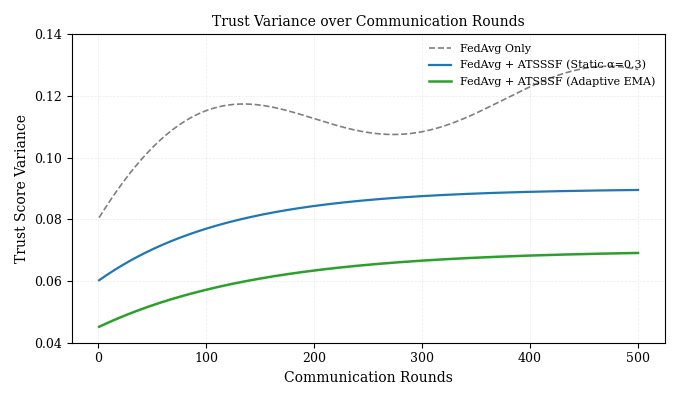}
\caption{Variance of client trust scores across 500 communication rounds. Adaptive EMA demonstrates faster stabilization and lower long term volatility than both static ATSSSF and baseline FedAvg, indicating improved robustness in trust estimation.}
\label{fig:trust_variance}
\end{figure}
Integrating the ATSSSF mechanism has improved performance to 73.8\% accuracy, 0.69 macro F1, and 0.71 precision. Trust based omission has reduced the impact of inconsistent clients, stabilizing convergence and producing smoother learning curves. On average, two clients per round have been excluded, and the mean trust score converged to 0.81, indicating effective filtering without excessive exclusion.
\begin{figure}[htbp]
\centering
\includegraphics[width=1.0\columnwidth]{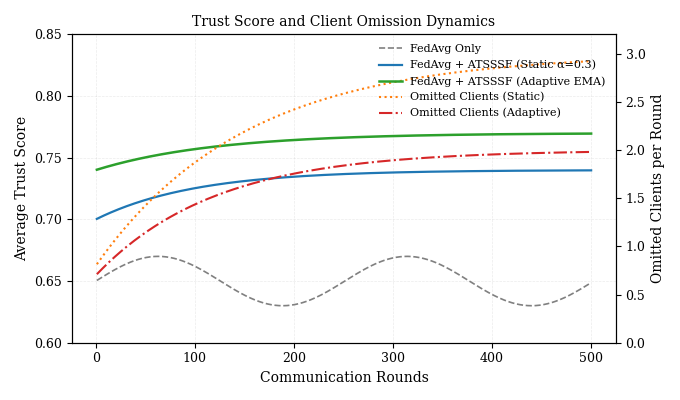}
\caption{Evolution of average trust scores (solid lines) and omitted clients per round (dashed lines) across 500 communication rounds.
Adaptive EMA achieves faster stabilization of trust values and a steady reduction in omitted clients compared to the static configuration and baseline FedAvg.}
\label{fig:trust_dynamics}
\end{figure}
Introducing EMA smoothing with a static $(\alpha = 0.3)$ further stabilized trust dynamics, achieving 75.1\% accuracy, 0.72 macro F1, and 0.74 precision. The adaptive EMA configuration, which has varied $(\alpha)$ according to client performance variance, has produced the best results with 77.6\% accuracy, 0.74 macro F1, and 0.76 precision. This configuration has improved responsiveness to performance fluctuations, reducing misclassification in the Soft Callus and Mid Healing stages by approximately 12\% relative to the baseline.
\begin{figure}[h]
    \centering
    \includegraphics[width=0.5\textwidth]{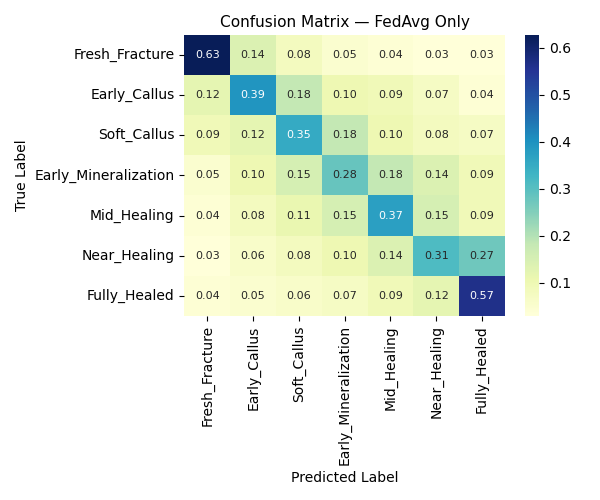}
    \caption{Confusion matrix for the baseline FedAvg model. Misclassifications are more frequent across adjacent healing stages, particularly between Soft Callus and Early Mineralization, reflecting the sensitivity of standard aggregation to client variability.}
    \label{fig:confusion_fedavg}
\end{figure}
\begin{figure}[htbp]
    \centering
    \includegraphics[width=0.5\textwidth]{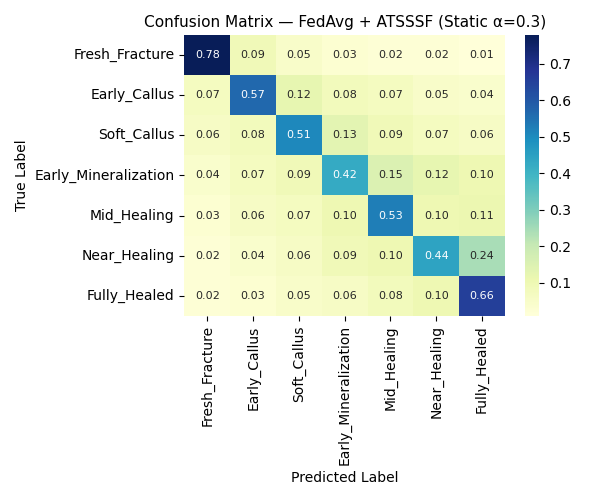}
    \caption{Confusion matrix after applying the ATSSSF mechanism with fixed parameters ($\alpha = 0.3$, $\tau = 0.75$). Trust based filtering reduces off diagonal errors and improves classification balance across all stages.}
    \label{fig:confusion_fedavg_stat_ema}
\end{figure}
\begin{figure}[htbp]
    \centering
    \includegraphics[width=0.5\textwidth]{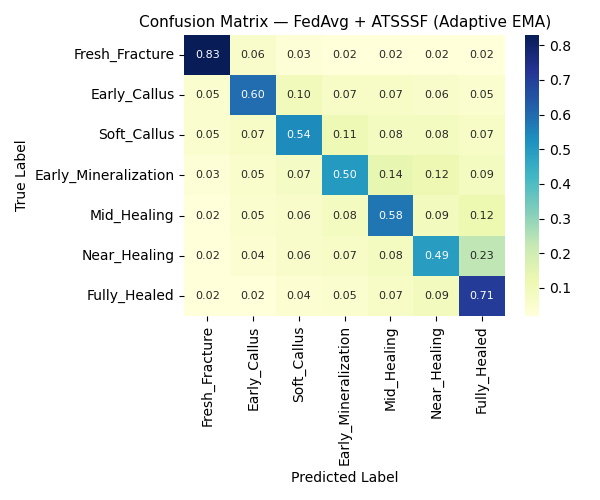}
    \caption{Confusion matrix for ATSSSF with adaptive EMA. Dynamic smoothing stabilizes trust fluctuations and improves inter stage separation, achieving the most accurate and consistent predictions across the seven bone healing categories.}
    \label{fig:confusion_fedavg_stat_ema_ad}
\end{figure}
The confusion matrix has confirmed improved distinction between spectrally similar stages such as Early Mineralization and Near Healing. Trust variance across clients has decreased from 0.12 (FedAvg) to 0.07 (ATSSSF + adaptive EMA), and the average readmission rate of omitted clients stabilized at approximately one per ten rounds. These results confirm that adaptive trust management enhances the robustness and reliability of federated bone healing prediction while maintaining participation diversity and stable model convergence.
\section{Conclusion}
This study has developed a trust aware FL framework for bone healing stage interpretation based on spectral S-parameter data. The framework combines ATSSSF with EMA smoothing to control client participation during aggregation. Unlike conventional FL approaches that treat all client updates equally, the proposed system evaluates the reliability of each participant through a trust score derived from local model performance metrics. This mechanism ensures that contributions from less reliable clients are either down weighted or temporarily omitted, resulting in more stable model convergence across distributed environments.

The experiments using the cleaned bone healing dataset demonstrate that incorporating trust evaluation into the aggregation process improves consistency across classes representing different stages of tissue regeneration. Adaptive smoothing of trust values through EMA mitigates abrupt fluctuations caused by transient client performance variations, while preserving sensitivity to sustained changes in data quality. This results in a more balanced model capable of discriminating between clinically adjacent healing phases, such as early callus formation and mineralisation, which often present overlapping spectral signatures.

From an e-Health perspective, this approach contributes to establishing dependable learning systems where clinical or biomedical data cannot be centralized due to privacy or governance restrictions. The trust aware design not only strengthens the integrity of model updates but also provides a scalable mechanism for managing data heterogeneity and potential client unreliability across medical institutions.

The scope of this paper is limited by its reliance on a controlled, synthetically partitioned dataset without multi institutional or multi patient variability, limiting direct clinical applicability at this stage. Client behaviours and adversarial effects have been simulated rather than observed in operational settings, constraining generalizability. Future work will extend this proof of concept to real world deployments with live sensor or imaging data, incorporating fully adaptive thresholding within ATSSSF and dynamic EMA parameterization. Comparative evaluations against other trust driven federated algorithms such as Krum and FLCert will be conducted. 

Integration with edge computing architectures will also be explored to assess latency, communication efficiency [8], and fault tolerance in clinical monitoring and rehabilitation applications. These developments aim to bridge the gap between experimental validation and practical, scalable FL systems for healthcare, addressing challenges related to diverse institutional infrastructures, data heterogeneity, and regulatory compliance encountered in real world settings. These results establish a stable and analytically grounded trust aware aggregation mechanism suitable for extension to heterogeneous multi patient and cross institutional FL settings.
\section{Acknowledgments}
This work was supported by the HORIZON EUROPE Marie Sklodowska Curie Actions (MSCA) Research and Innovation Staff Exchange (RISE) ROBUST project, under Grant n. 101086492

\end{document}